\documentclass[letterpaper]{article}
\usepackage[utf8]{inputenc}
\usepackage[preprint]{aaai2027}
\usepackage[hyphens]{url}
\usepackage{graphicx}
\usepackage{natbib}
\usepackage{caption}
\usepackage{amsmath}
\usepackage{amssymb}
\usepackage{algorithm}
\usepackage{algorithmic}
\usepackage{hyperref}
\usepackage{newfloat}
\usepackage{listings}
\DeclareCaptionStyle{ruled}{labelfont=normalfont,labelsep=colon,strut=off}
\floatstyle{ruled}
\newfloat{listing}{tb}{lst}{}
\floatname{listing}{Listing}

\usepackage{booktabs}
\usepackage{array}
\usepackage{tabularx}
\usepackage{multirow}

\title{AppDeltaWorld: Transition-Grounded Delta Code World Model for Mobile GUI Agents}
\author{
    \textbf{Weikai Xu$^{1*}$, Yunren Feng$^{2*}$, Haoxiang Lei$^{2*}$, Kun Huang$^{3\dagger}$, Yuxuan Liu$^{4}$} \\
    \textbf{Kang Zhao$^{3\dagger}$, Xiaolin Hu$^{5}$, Shuo Shang$^{2\ddagger}$, Bo An$^{1\ddagger}$} 
    }

\affiliations{
$^1$\textmd{Nanyang Technological University, Singapore} 
  $^2$\textmd{University of Electronic Science and Technology of China} \\
  $^3$\textmd{Independent Researchers} 
  $^4$\textmd{Gaoling School of Artificial Intelligence, Renmin University of China} 
   \\
   $^5$\textmd{Xiamen University} \\
   $^*$\textmd{ Equal contribution.}
   $\dagger$\textmd{ Project Leader.}
   $\ddagger$\textmd{ Corresponding authors.} \\}

\begin{document}

\maketitle

\begin{abstract}
Mobile GUI agents can operate apps through pixel perception and touch actions, making them a promising interface for collecting and improving long-horizon mobile interaction policies.
However, real trajectories are difficult to obtain for sensitive apps and privacy-critical operations. At the same time, existing simulated environments are costly to scale up, and GUI world models still suffer from unstable generation, limited modality coverage, and inconsistent action-transition logic.
To address these limitations, we propose AppDeltaWorld, a transition-grounded delta code world model that predicts the next GUI as a reachable code update rather than as an unconstrained image or text description.
AppDeltaWorld retrieves app-specific Level-1 HTML references under an action-transition constraint, generates Level-2 executable HTML conditioned on the current screen, action, predicted next-screen text, and retrieved structure, and inserts generated visual assets into image slots before browser rendering.
As a world model, AppDeltaWorld achieves the highest fidelity on CMGUIBench-500 under Code2World evaluation, with clear gains in structural layout and UI element reconstruction over image-only and code-only baselines.
As a training environment, AppDeltaWorld supports filtered closed-loop SFT data construction that, when combined with public supervision, enables AppDeltaAgent to achieve state-of-the-art performance on AndroidLens and consistent gains on MobileGym and MobileWorld. 
Moreover, world-model-based test-time reinforcement learning enables policy adaptation and shows further improvements without additional interaction with real apps. 

\end{abstract}

\section{Introduction}
Recent advances in multimodal foundation models and increasingly capable agent interaction paradigms have renewed interest in mobile GUI agents~\citep{lu2026ui,yang2026gui,qin2025ui}. Unlike API- or CLI-based agents~\citep{xiao2026webworld}, which can only act through interfaces exposed by apps and are constrained by platform authorization and permission boundaries, mobile GUI agents can operate smartphones directly through pixel perception and touch actions. This general interaction makes them broadly applicable across heterogeneous mobile apps. Still, it also faces a data access bottleneck: sensitive operations and privacy-critical apps often lack real interactive trajectories. Existing works have therefore explored two solutions: building executable simulated environments~\citep{tang2026phoneworld,dong2026agent}, or training GUI world models that can render next states~\citep{zuo2026qwen}, enabling closed-loop agent optimization without these private data.

\begin{figure}[t]
    \centering
    \includegraphics[width=\columnwidth]{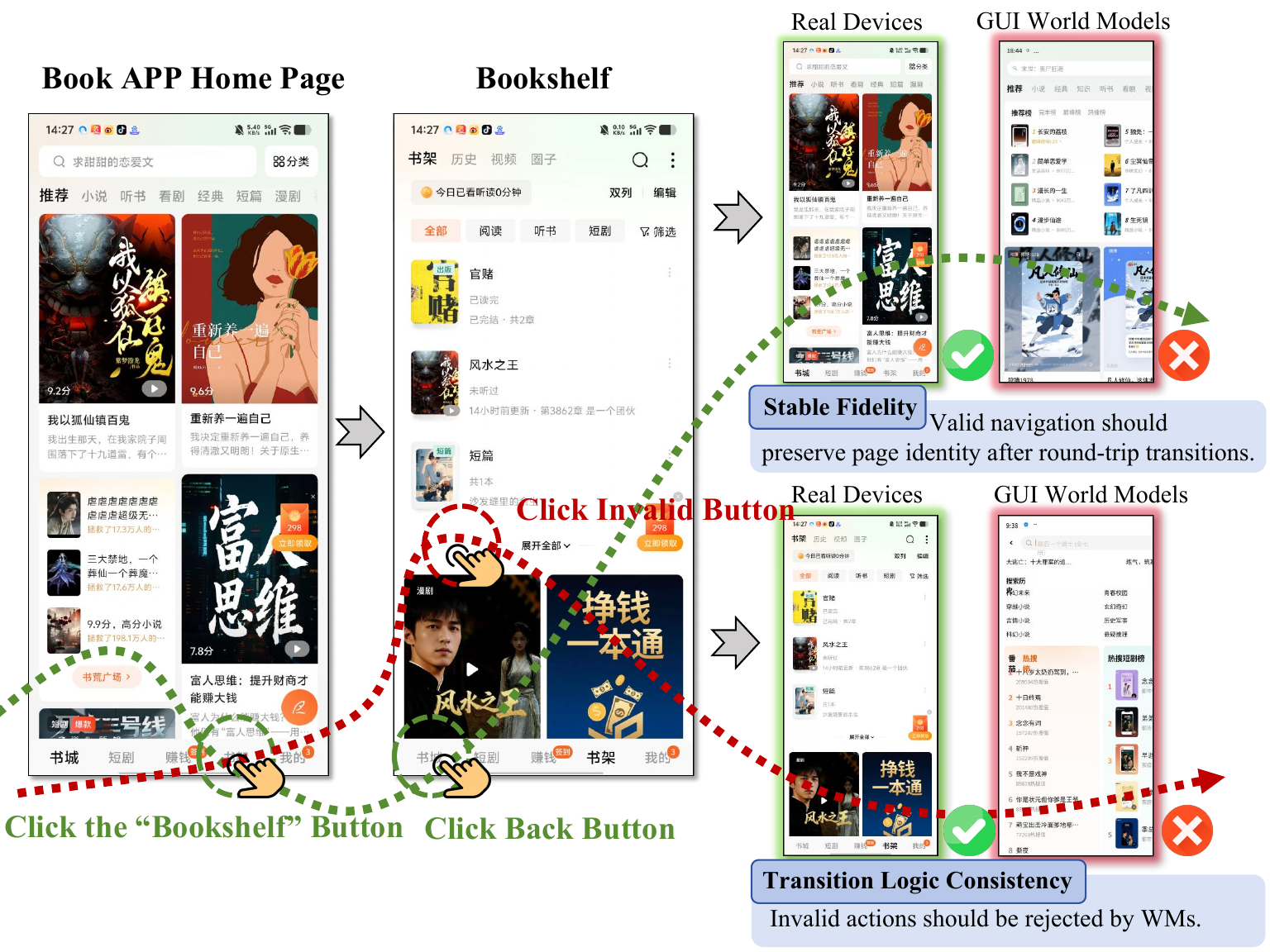}
    \vspace{-0.3cm}
    \caption{Poor fidelity (upper) and inconsistent action interaction logic (lower) problems in GUI world models. }
    \label{fig:motivation}
    \vspace{-0.3cm}
\end{figure}
However, scaling either approach to broad app coverage remains challenging. Simulator-based environments require substantial manual effort to expand app coverage and implement app-specific functionality, making them expensive to maintain as mobile interfaces evolve. GUI world models reduce this engineering burden, but their usefulness is still limited by the following three requirements. (1) \textbf{Stable Fidelity}: as shown in Figure~\ref{fig:motivation}, when the agent performs complex actions during trajectory collection, world models should be able to stably output rendered pages that are highly consistent with real devices. (2) \textbf{Hybridization Modality}: a text-only world model cannot directly provide visual training data. In contrast, an image-only model often fails to represent dense UI text and fine-grained layout. (3) \textbf{Transition Logic Consistency}: the predicted transition must follow the interaction logic in real apps, which requires world models to express a refusal when faced with invalid actions.

To address these problems, we propose \textbf{AppDeltaWorld}, a transition-grounded \textbf{delta} code \textbf{world} model for complex mobile \textbf{apps}. It explicitly separates structural retrieval, multimodal rendering, and action-transition validation. First, for stable fidelity, we decompose HTML code into two levels: a Level-1 HTML state captures reusable page structure and layout, while a Level-2 HTML code completes the concrete next screen. AppDeltaWorld first localizes the current screen to a source cluster, uses an app-specific transition memory to identify action-reachable target clusters, and retrieves the Level-1 code from these reachable candidates. Then, the Level-2 HTML code is generated conditioned on this structural reference, avoiding the need to regenerate the whole interface from scratch and making repeated visits to similar pages more consistent. Second, for hybrid modality, AppDeltaWorld keeps dense UI text and geometry in executable HTML, but delegates visual regions that are difficult to express with code to image synthesis. Specifically, the generated Level-2 HTML contains textual descriptions for image slots, and a text-to-image model inserts the corresponding visual assets before browser rendering. This combines the precision of code for layout and text with the visual richness of generative images. Third, for transition logic consistency, AppDeltaWorld builds an action-transition index for each app. Before selecting a next-screen reference, it constrains target retrieval by the source cluster and action target, so unsupported actions are treated as invalid or low-confidence transitions rather than unconstrained hallucinated states. 

\begin{figure*}[t]
  \centering
  \includegraphics[width=\textwidth]{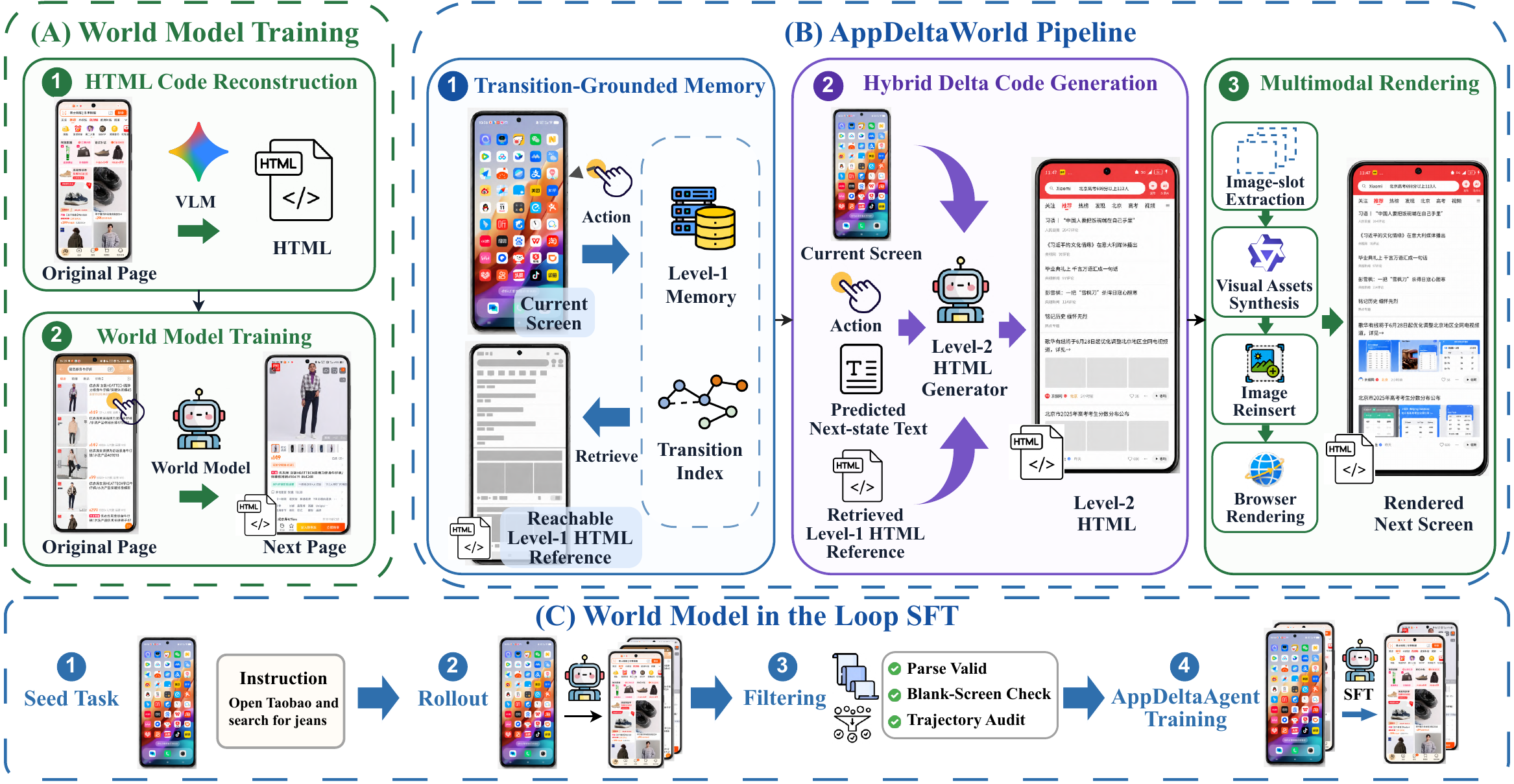}
  \vspace{-0.5cm}
  \caption{\textbf{Overview of AppDeltaWorld}: (A) original mobile pages are reconstructed as HTML to train the world model; (B) transition-grounded memory retrieves an action-reachable Level-1 HTML reference that guides hybrid delta code generation and multimodal rendering of the next GUI; and (C) filtered closed-loop rollouts are converted into SFT data for AppDeltaAgent.}
  \label{fig:appdeltaworld}
  \vspace{-0.5cm}
\end{figure*}
We evaluate AppDeltaWorld as both a world model and a data-generation environment for GUI agents. For world-model learning, we build a transition-centric training mixture with 100,149 GUI transition steps, dominated by CMGUI and complemented by CAGUI, Magic-RICH, and ChiM-Nav. With this data and the transition-grounded RAG pipeline, AppDeltaWorld achieves the best overall score on CMGUIBench-500 under the Code2World evaluation, surpassing GPT-Image-2 and Gemini-3.1-Pro-Image. Using GUI-Owl and OpenMobile as rollout seeds, we generate 33,133 AppDelta trajectories with AppDeltaWorld as expanding data to train \textbf{AppDeltaAgent}. The generated data is evaluated by downstream performance: AppDeltaAgent achieves the SoTA performance on AndroidLens. In online evaluation, it improves over the Qwen3-VL-8B base on MobileGym from 10.2\% to 14.1\% SR, and on MobileWorld from 9.4\% to 14.9\% GUI-only SR. Additional test-time app-specific RL training also shows consistent improvements, indicating that AppDeltaWorld can support both supervised rollout construction and subsequent policy optimization.

Our contributions are summarized as follows:
\begin{itemize}
    \item We introduce transition-grounded hierarchical HTML code RAG, which retrieves Level-1 reference code and uses app-specific action transitions to constrain Level-2 next-screen generation.
    \item We design a hybrid text-code-diffusion world model that combines semantic retrieval, executable HTML generation, and image-slot rendering.
    \item We validate AppDeltaWorld on CMGUIBench-500 and show that world-model-in-the-loop SFT and RL-training improve downstream AppDeltaAgent evaluation.
\end{itemize}

\section{Related Work}

\subsection{Mobile GUI Agents}
Foundation models shift mobile GUI agents from metadata-driven automation to direct visual interaction~\citep{deng2024mobile,xusman}. Earlier systems relied on incomplete or inconsistent view hierarchies and accessibility trees, while multimodal LLMs enable screenshot-based perception~\citep{shen2024falcon}. Recent agents use screenshot-based plan-and-act pipelines, including AppAgent, CogAgent, and MobileSteward~\citep{zhang2025appagent,hong2023cogagent,liu2025mobilesteward}, and Mobile-Agent adds multi-agent collaboration for long-horizon execution~\citep{wang2024mobile,wang2025mobile}. Other work improves screen understanding, reasoning, grounding, and policy learning through mid-training or reinforcement learning~\citep{huang2025mobileipl,liu2026come,wang2025ui,xu2025mobile,shi2025mobilegui,chen2025step}. However, most agents remain reactive, selecting actions from current observations without anticipating future states, which can amplify navigation errors~\citep{zhang2024android}.

\subsection{GUI World Models}
World models predict future states from observations and actions, and recent vision-language models extend them to GUI automation~\citep{gu2024your,chae2024web,guan2026computer,jiang2026r,sun2024determlr}. Existing GUI world models use text-, pixel-, or code-level representations. Text-level methods model GUI transitions through semantic descriptions~\citep{li2025word}. MobileWorld predicts future states with natural-language descriptions and question-answer pairs~\citep{li2025mobileworldbench}, whereas MobileDreamer preserves coarse layouts through task-relevant text sketches~\citep{cao2026mobiledreamer}. Pixel-level methods synthesize future screenshots; ViMo uses diffusion models but is costly and struggles with dense text and small UI elements~\citep{luo2025vimo}. Code-level methods, including Code2World and gWorld, generate executable UI code and render screenshots with browser engines, using symbolic intermediate representations to improve visual fidelity~\citep{zheng2026code2world,koh2026generative,xu2026mobile}. Existing work mainly treats GUI world models as next-state predictors, leaving their use in mobile-agent decision making and training underexplored.

\section{Methodology}

\subsection{Transition-Grounded Level-1 Code Retrieval}\label{sec:Transition-Grounded Code Retrieval}

Code-based GUI world models require stability and transition consistency during long-horizon rollouts. 
Generating the next HTML directly from the current screenshot and action risks structural drift or predicting an unreachable target state in the real app. AppDeltaWorld addresses this by turning it into a transition-grounded retrieval-and-completion problem: it first localizes the current screen to an app-specific source state, then constrains target retrieval to action-reachable screen families, and finally retrieves a Level-1 HTML reference before generating the Level-2 HTML code.

Given a current screenshot $x_t$, an app identifier $a$, and a mobile action $m_t$, AppDeltaWorld first converts the current screen into a structured textual state $r_t$, which can be represented as
\begin{equation}
 r_t = \phi_{\mathrm{text}}(x_t)
     = [f_t, d_t, \ell_t, q_t, p_t]
\end{equation}
where $f_t$, $d_t$, $\ell_t$, $q_t$, and $p_t$ denote the functional template, screen description, region layout, slot schema, and page type, respectively. This current-state representation is used to identify the source screen family, while the final reference HTML is selected by a separate target-side retrieval query. As shown in Figure~\ref{fig:appdeltaworld}(A), each historical screen is reconstructed as executable HTML to provide supervision for world-model training and stored in the memory index as a tuple $(r_i, h_i^{(1)}, c_i^{(0)}, c_i^{(1)})$, where $r_i$ is the retrieval text, $h_i^{(1)}$ is the Level-1 reference HTML, $c_i^{(0)}$ is its coarse functional category, and $c_i^{(1)}$ is its fine-grained Level-1 cluster. The offline index first assigns each screen to a coarse functional category using a rule-based classifier:
\begin{equation}
c_i^{(0)} = g(p_i, f_i, \ell_i, q_i)
\end{equation}
where $g(\cdot)$ maps the page type, functional template, region layout, and slot schema to a predefined functional category. It then builds separate hashed TF--IDF vectors for layout, DOM signature, semantic text, and slot schema and combines them into a weighted representation:
\begin{equation}
z_i = 0.45z_i^{\mathrm{layout}} + 0.25z_i^{\mathrm{dom}}
    + 0.20z_i^{\mathrm{sem}} + 0.10z_i^{\mathrm{slot}}
\end{equation}
Within each coarse functional category $c_i^{(0)}$, cosine-similarity clustering over $z_i$ produces the fine-grained cluster $c_i^{(1)}$. Thus, $c_i^{(1)}$ is nested under $c_i^{(0)}$. 
Test-time retrieval and transition constraints operate directly on $c_i^{(1)}$, so $c_i^{(0)}$ is not queried separately during inference.
AppDeltaWorld forms a source query $q_t^{\mathrm{src}}=[a,u,r_t,m_t]$ from the app, instruction, current state, and action, and localizes a source cluster by
\begin{equation}
 i_t^{\mathrm{src}}
 = \arg\max_{i:a_i=a}
   \langle e(q_t^{\mathrm{src}}), e(r_i) \rangle,
 \qquad
 c_t = c_{i_t^{\mathrm{src}}}^{(1)} 
\end{equation}
where $u$ denotes the user instruction and $e(\cdot)$ is the retrieval vectorizer. 
Since the source decision is made at the cluster level, repeated visits to the same functional screen are localized to the same source cluster, allowing world models to consider only the incremental part.

To preserve action logic, AppDeltaWorld constrains target retrieval with a transition index before selecting a reference HTML. For each source cluster $c_t$ and action $m_t$, the transition memory stores candidate target clusters. For click and long-press actions, the action target is quantized into a $6\times 12$ grid; for swipe actions, both start and end grids can be used. Let $\kappa(m_t)$ be the action-target key and let $\mathcal{T}$ be the transition index. The allowed target set is
\begin{equation}
 \mathcal{C}_{t+1} = \mathcal{T}(c_t, \alpha(m_t), \kappa(m_t))
\end{equation}
where $\alpha(m_t)$ is the action type. 
AppDeltaWorld then predicts a semantic next-screen state $\hat{s}_{t+1}$ and forms a target query $q_{t+1}^{\mathrm{tar}}=[a,u,m_t,\hat{s}_{t+1}]$. The final Level-1 reference is selected by constrained retrieval:
\begin{equation}
 i^*
 = \arg\max_{i:a_i=a,\; c_i^{(1)} \in \mathcal{C}_{t+1}}
   \langle e(q_{t+1}^{\mathrm{tar}}), e(r_i) \rangle 
\end{equation}
The retrieved $h_{i^*}^{(1)}$ provides a reachable structural basis for next-screen generation, while transitions without supported target clusters are rejected as invalid or low-confidence.

\begin{algorithm}[t]
\footnotesize
\caption{World-model-in-the-loop rollout construction.}
\label{alg:wm-sft-rollout}
\begin{algorithmic}[1]
\setlength{\itemsep}{0pt}
\setlength{\parsep}{0pt}
\setlength{\parskip}{0pt}
\REQUIRE Number of seeds $N$, horizon $T$, policy $\pi_\eta$, world model $\mathcal{W}$, and evaluator $\mathcal{E}$
\ENSURE $\mathcal{D}_{\mathrm{sft}}$ and $\mathcal{Q}$
\STATE $\Xi\leftarrow\textsc{Seed}(N),\quad \mathcal{D}_{\mathrm{sft}},\mathcal{Q}\leftarrow\emptyset,\emptyset$
\FOR{$(x_0,u,a)\in\Xi$}
  \STATE $h_0\leftarrow\emptyset,\ \tau\leftarrow[x_0]$
  \FOR{$t=0,\ldots,T-1$}
    \STATE $(v_t,m_t)\leftarrow\textsc{Parse}\!\left(\pi_\eta(x_t,u,h_t)\right)$ \hfill {\scriptsize\textit{// predict an action}}
    \STATE \textbf{if} $v_t=0$ \textbf{then break}
    \STATE \textbf{if} $m_t\in\mathcal{M}_{\mathrm{term}}$ \textbf{then} $\mathcal{D}_{\mathrm{sft}}\leftarrow\mathcal{D}_{\mathrm{sft}}\cup\{(x_t,u,h_t,m_t)\}$; \textbf{break}
    \STATE $(x_{t+1},\mu_t)\leftarrow\mathcal{W}(x_t,m_t,u,a)$ \hfill {\scriptsize\textit{// generate the next GUI state}}
    \STATE \textbf{if} $\textsc{QC}(x_{t+1},\mu_t)=0$ \textbf{then break} \hfill {\scriptsize\textit{// reject invalid transitions}}
    \STATE $\mathcal{D}_{\mathrm{sft}}\leftarrow\mathcal{D}_{\mathrm{sft}}\cup\{(x_t,u,h_t,m_t)\}$
    \STATE $h_{t+1}\leftarrow h_t\cup\{m_t\},\quad \tau\leftarrow\tau\cup\{(m_t,x_{t+1})\}$
  \ENDFOR
  \STATE $\mathcal{Q}\leftarrow\mathcal{Q}\cup\{\mathcal{E}(\tau)\}$
\ENDFOR
\RETURN $\mathcal{D}_{\mathrm{sft}},\mathcal{Q}$
\end{algorithmic}
\end{algorithm}

\subsection{Hybrid Multimodal Rendering Framework}

AppDeltaWorld combines text for semantic retrieval and transition grounding, code for executable layout, and diffusion-based synthesis for visual slots (Figure~\ref{fig:appdeltaworld}(B)). Accordingly, next-state prediction comprises three stages:
\begin{equation}
 \hat{x}_{t+1}
 = \mathcal{R}\!\left(
    \mathcal{I}\!\left(
      \mathcal{G}_{\theta}(x_t, m_t, \hat{s}_{t+1}, h_{i^*}^{(1)})
    \right)
   \right)
\end{equation}
where $\hat{s}_{t+1}$ is the predicted next-screen text, $\mathcal{G}_{\theta}$ is the multimodal HTML world model, $\mathcal{I}$ inserts generated illustration assets, and $\mathcal{R}$ renders the Level-2 HTML into a screenshot.
The text stage predicts $\hat{s}_{t+1}$ as the target-side semantic query for Level-1 HTML retrieval under the transition-constrained cluster set introduced in the last subsection. By separating semantic next-state prediction from HTML synthesis, the world model can leverage text for tasks at which it is most effective: matching screen intent, page type, visible entities, and expected state changes.
Then the code stage generates a Level-2 HTML screen as a delta from the retrieved Level-1 reference. The multimodal generator receives the current screenshot, the structured action record, the predicted next-screen text, and the retrieved reference HTML:
\begin{equation}
 \hat{h}_{t+1}^{(2)} = \mathcal{G}_{\theta}(x_t, m_t, \hat{s}_{t+1}, h_{i^*}^{(1)})
\end{equation}
Here, $h_{i^*}^{(1)}$ supplies reusable app-specific layout, widgets, and DOM organization, while $x_t$, $m_t$, and $\hat{s}_{t+1}$ determine how this structure should change after the action. This delta code formulation directly addresses the instability of free-form screen generation, especially when there are multiple similar candidate states.
The visual stage augments the generated HTML with visual assets for image-rich regions that are difficult to capture through code alone. Qwen-Image-style text-to-image synthesis fills image slots using textual descriptions extracted from the HTML, and the resulting HTML is rendered by a browser-based renderer. On product pages and in video applications, the absence of visual assets in image slots reduces the prediction fidelity of the world model when these slots occupy the majority of the page. These observations indicate that the involvement of a diffusion model is crucial for constructing training trajectories.

\begin{table}[t]
\footnotesize
\renewcommand{\arraystretch}{0.75}
\centering
\vspace{-0.3cm}
\caption{Training data for the world model and action model.}
\label{tab:training-data-mixture}
\scriptsize
\setlength{\tabcolsep}{4.0pt}
\resizebox{0.9\columnwidth}{!}{%
\begin{tabular}{l r c}
\toprule
\textbf{Source} & \textbf{Steps / Samples} & \textbf{Ratio} \\
\midrule
\multicolumn{3}{c}{\textit{World-model training data}} \\
\midrule
CMGUI~\citep{xie2026secagent} & 95,614 & 95.47\% \\
CAGUI~\citep{zhang2025agentcpm} & 2,978 & 2.97\% \\
Magic-RICH~\citep{tang2025magicgui} & 1,304 & 1.30\% \\
ChiM-Nav~\citep{zhang2026omegause} & 253 & 0.25\% \\
\midrule
\multicolumn{3}{c}{\textit{Action-model training data}} \\
\midrule
GUI-Owl (10\% Sampled)~\citep{gao2026guitester} & 56,237 & 48.18\% \\
AppDelta (Constructed by AppDeltaWorld) & 33,133 & 28.38\% \\
OpenMobile~\citep{cheng2026openmobile} & 27,360 & 23.44\% \\
\bottomrule
\end{tabular}%
}
\vspace{-0.2cm}
\end{table}

\subsection{World-Model-in-the-Loop SFT Data Construction}

Unlike previous work, we are more concerned about whether AppDeltaWorld can serve as a substitute for the real environment, providing learnable experience for action policy models.
Algorithm~\ref{alg:wm-sft-rollout} and Figure~\ref{fig:appdeltaworld}(C) summarize this rollout construction and filtering procedure. We use $N$ rollout seeds from GUI-Owl and OpenMobile with a maximum horizon $T$. The action policy $\pi_\eta$ predicts a raw mobile-use response $\tilde{m}_t$, which is parsed into a validity flag $v_t$ and a structured action $m_t$. The world model $\mathcal{W}$ denotes the full transition-grounded generation pipeline, including retrieval, HTML generation, image insertion, and rendering; it returns the next rendered observation $x_{t+1}$ and metadata $\mu_t$ such as retrieval records and render paths. The evaluator $\mathcal{E}$ records trajectory-level quality in $\mathcal{Q}$, while $\mathcal{D}_{\mathrm{sft}}$ stores accepted SFT trajectories.

\begin{table*}[t]
\renewcommand{\arraystretch}{0.75}
\centering
\footnotesize
\vspace{-0.5cm}
\caption{\textbf{Main results on CMGUIBench-500.} We report the functional-logic scores ($S_{ad}$ and $S_{id}$) judged by Gemini-3-Flash, visual-quality scores ($S_{ele}$, $S_{lay}$, SigLIP, and DINOv2), the average number of output tokens when available, and the overall score.}
\label{tab:cmguibench500-code2world}
\scriptsize
\setlength{\tabcolsep}{3.2pt}
\resizebox{0.9\textwidth}{!}{%
\begin{tabular}{l l c c c c c c c c}
\toprule
\multirow{2}{*}{\textbf{Model}} & \multirow{2}{*}{\textbf{Modality}} &
\multicolumn{2}{c}{\textbf{Functional Logic}} &
\multicolumn{4}{c}{\textbf{Visual Quality}} &
\multirow{2}{*}{\textbf{Avg. Tokens}} & \multirow{2}{*}{\textbf{Overall}} \\
\cmidrule(lr){3-4} \cmidrule(lr){5-8}
& & $S_{ad}$ & $S_{id}$ & $S_{ele}$ & $S_{lay}$ & \textit{SigLIP} & \textit{DINOv2} & & \\
\midrule
\multicolumn{10}{c}{\textit{Image Generation}} \\
\midrule
    GPT-Image-2 & Diffusion & \textbf{91.73} & \underline{83.40} & \underline{44.00} & \underline{41.70} & \underline{93.15} & \textbf{80.07} & -- & \underline{72.34} \\
    Gemini-3.1-Pro-Image & Diffusion & 88.53 & \textbf{85.00} & 42.18 & 40.18 & \textbf{93.43} & \underline{80.07} & -- & 71.57 \\
    Gemini-3-Flash-Image & Diffusion & 87.67 & 84.39 & 40.28 & 41.32 & 90.64 & 78.89 & -- & 70.53 \\
\midrule
\multicolumn{10}{c}{\textit{Code Generation}} \\
\midrule
    GPT-5.4 & Code & 82.17 & 74.40 & 35.46 & 33.77 & 90.63 & 75.64 & 4960.50 & 65.35 \\
    Gemini-3-Flash & Code & \underline{90.85} & 83.20 & 36.93 & 36.13 & 82.08 & 58.31 & 1818.40 & 64.58 \\
    Mimo-v2.5 & Code & 89.83 & 81.80 & 33.69 & 32.21 & 85.34 & 64.19 & 2730.70 & 64.51 \\
    Claude-Sonnet-4.6 & Code & 87.00 & 79.80 & 36.00 & 34.58 & 83.00 & 60.26 & 2705.10 & 63.44 \\
    Qwen3.7-Max & Code & 72.55 & 69.60 & 27.67 & 26.36 & 84.04 & 62.54 & 3355.70 & 57.13 \\
\midrule
    Qwen3.5-27B & Code & 89.63 & 82.20 & 32.59 & 31.01 & 82.73 & 60.38 & 3455.10 & 63.09 \\
    Qwen3.6-27B & Code & 85.36 & 77.40 & 32.73 & 31.83 & 82.35 & 59.79 & 4109.80 & 61.58 \\
    MiniMax-M3 & Code & 76.46 & 69.40 & 30.88 & 29.24 & 86.46 & 67.02 & 2759.80 & 59.91 \\
    Qwen3.5-9B & Code & 80.04 & 75.60 & 27.06 & 25.22 & 81.72 & 59.74 & 3525.00 & 58.23 \\
    Kimi-K2.6 & Code & 70.70 & 62.60 & 27.76 & 26.49 & 83.37 & 62.37 & 4379.10 & 55.55 \\
    Qwen3-8B & Code & 43.80 & 46.20 & 28.17 & 26.23 & 87.33 & 71.42 & 4157.00 & 50.53 \\
    InternVL3.5-38B & Code & 42.22 & 43.60 & 23.04 & 20.93 & 79.78 & 57.21 & 850.70 & 44.46 \\
    Code2World-8B & Code & 63.23 & 66.51 & 37.02 & 36.59 & 81.21 & 55.30 & 4457.60 & 56.64 \\
\midrule
\multicolumn{10}{c}{\textit{Hybrid Multimodal Generation}} \\
\midrule
ViMo & Text+Diffusion & 42.44 & 42.40 & 14.18 & 13.22 & 85.20 & 63.37 & -- & 43.44 \\
\textbf{AppDeltaWorld (Ours)} & Text+Code+Diffusion & 79.69 & 77.00 & \textbf{56.26} & \textbf{57.43} & 91.84 & 78.82 & 8309.40 & \textbf{73.51} \\
\bottomrule
\end{tabular}%
}
\vspace{-0.3cm}
\end{table*}
\renewcommand{\arraystretch}{1.0}

\begin{figure}[t]
    \centering
    \begin{minipage}{0.485\columnwidth}
        \centering
        \includegraphics[width=\linewidth]{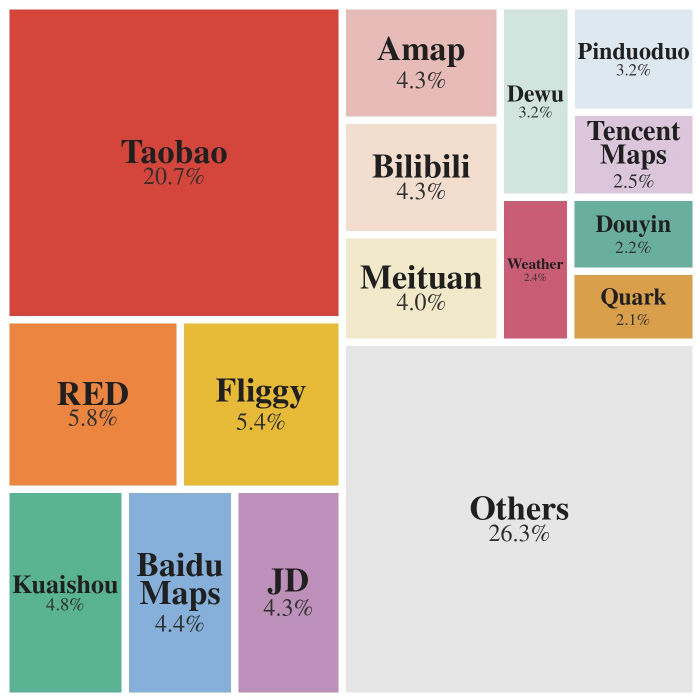}\\[-1mm]
        {\scriptsize (a) World model}
    \end{minipage}\hfill
    \begin{minipage}{0.485\columnwidth}
        \centering
        \includegraphics[width=\linewidth]{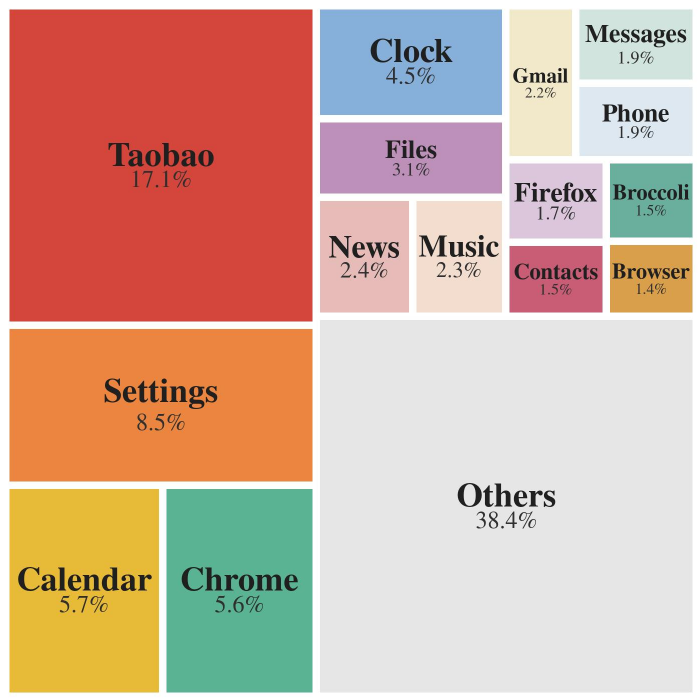}\\[-1mm]
        {\scriptsize (b) Action model}
    \end{minipage}
    \vspace{-0.3cm}
    \caption{Training APP distributions. The left side is AppDeltaWorld; the right side is AppDeltaAgent.}
    \label{fig:training-app-distribution}
    \vspace{-0.5cm}
\end{figure}

This data construction pipeline starts from real task seeds rather than synthetic prompts alone. We sample initial screenshots and instructions from GUI-Owl and OpenMobile, store them as rollout seeds, and then run an action model inside the AppDeltaWorld environment. Formally, a seed is $\xi=(x_0,u,a)$, where $x_0$ is the initial screenshot, $u$ is the user instruction, and $a$ is the app metadata. At step $t$, the action model predicts a mobile tool call from the current rendered observation and the action history:
\begin{equation}
 m_t \sim \pi_{\eta}(\cdot \mid x_t, u, h_t),
 \qquad
 h_t=(m_0,\ldots,m_{t-1})
\end{equation}
If $m_t$ is a termination action, the trajectory stops. Otherwise, the action is converted into the same pair format used by the world-model pipeline, and AppDeltaWorld predicts and renders the next observation:
\begin{equation}
 x_{t+1}=\mathcal{R}\!\left(\mathcal{I}\!\left(
   \mathcal{G}_{\theta}(x_t,m_t,\hat{s}_{t+1},h_{i^*}^{(1)})
 \right)\right)
\end{equation}
The rendered PNG is then fed back to the action model for the next step, producing a closed-loop synthetic trajectory $\tau=(x_0,m_0,x_1,\ldots,x_T)$. 

To prevent world-model artifacts from contaminating the supervision signal, we require that the action description policy and the action coordinate must be consistent; continuous repetitive actions are not accepted; and the final step must be an active termination rather than reaching the maximum steps. The accepted trajectory is defined as
\begin{equation}
\begin{aligned}
 \mathcal{D}_{\mathrm{sft}} = \{(x_t,u,h_t,m_t) \mid
 &\ \mathrm{parse}(m_t)=1, \\
 &\ \mathrm{blank}(x_{t+1})=0
    \ \text{or}\ m_t \in \mathcal{M}_{\mathrm{term}} \}
\end{aligned}
\end{equation}
An optional trajectory evaluator assigns completion and generation-quality scores for analysis or additional filtering. This procedure preserves transition-grounded rollout supervision while excluding trajectories corrupted by parsing or rendering failures.
\subsection{Training Data Statistics}
Table~\ref{tab:training-data-mixture} summarizes the final data sources used by the two training stages and Figure \ref{fig:training-app-distribution} shows the APP distribution. 
The world-model training data was all reverse engineered using Claude-4.8-Opus and Gemini-3.1-Pro to obtain renderable code; the code contains Level 1- and Level 2- labels to distinguish the structure and components.
When constructing supervision trajectories for action models, we try to ensure a balance among the APPs, but this is still affected by the distribution of upstream WM data. From the perspective of construction efficiency, only 1/10 of the data passed the quality verification. The main reasons for failure are: (1) \textbf{Task Progress Loss}, the world model fails to judge the progress of the task, resulting in multiple loops. (2) \textbf{Poor Simulation Fidelity}, the page quality drops significantly after the number of link steps increases. (3) the task is \textbf{Not Completed} within the specified number of steps. The differences in app data distribution will be discussed in subsequent experiments.

\section{Experiments}
\subsection{Settings}
\noindent \textbf{Benchmarks.} For world modeling, we use CMGUIBench-500 (500 random samples from the original CMGUI benchmark~\citep{xie2026secagent}) with Code2World evaluation. For action-policy evaluation, we report static action prediction results on AndroidLens~\citep{cao2026androidlens}, a task-level benchmark with both low-level and high-level instruction settings. We also include an online real environment, MobileWorld~\citep{kong2026mobileworld}, and a simulated environment, MobileGym~\citep{wu2026mobilegym}, to measure closed-loop task execution behavior.

\begin{table*}[!t]
\renewcommand{\arraystretch}{1.1}
  \centering
  \caption{\textbf{Results on AndroidLens.} HL denotes high-level instructions, LL denotes low-level instructions, AMS represents Action Matching Score, ATP represents Average Task Progress, and ``\# tokens" is the number of output tokens.}
  \small
  \begin{tabularx}{0.95\textwidth}{l*{12}{>{\centering\arraybackslash}X} >{\centering\arraybackslash}p{1.1cm}}
    \toprule
    \multirow{2}{*}{\textbf{Model}} 
    & \multicolumn{2}{c}{Chinese-LL} 
    & \multicolumn{2}{c}{Chinese-HL} 
    & \multicolumn{2}{c}{English-LL} 
    & \multicolumn{2}{c}{English-HL} 
    & \multicolumn{2}{c}{Total-LL}
    & \multicolumn{3}{c}{Total-HL} \\
    \cmidrule(r){2-3}
    \cmidrule(r){4-5}
    \cmidrule(r){6-7}
    \cmidrule(r){8-9}
    \cmidrule(r){10-11}
    \cmidrule(r){12-14}
    & AMS & ATP & AMS & ATP & AMS & ATP & AMS & ATP & AMS & ATP & AMS & ATP & {\# tokens} \\
    \hline
    \multicolumn{14}{c}{\textit{Agent workflows}} \\
    \hline
    Mobile-Agent-v2 & 47.18 & 23.38 & 32.66 & 13.53 & 47.46 & 23.14 & 33.43 & 13.38 & 47.27 & 23.21 & 32.92 & 13.46 & - \\
    Mobile-Agent-E & 74.52 & 38.27 & 50.12 & 20.81 & 75.22 & 37.54 & 49.85 & 19.97 & 74.98 & 37.65 & 50.03 & 20.07 & - \\
    GPT-4o + UGround-V1-7B & 64.15 & 31.64 & 46.08 & 21.59 & 66.63 & 35.46 & 44.68 & 20.18 & 69.73 & 38.44 & 45.61 & 21.16 & - \\
    AppAgent-v2 & 50.36 & 25.12 & 35.84 & 15.02 & 50.91 & 25.48 & 36.27 & 14.82 & 50.54 & 25.24 & 35.98 & 14.96 & 493.61 \\
    Mobile-Agent-v3.5 & 79.54 & 46.29 & 54.81 & 22.93 & 78.63 & 38.57 & 53.21 & 22.42 & 79.24 & 43.63 & 53.70 & 22.65 & 325.14 \\
    \hline
    \multicolumn{14}{c}{\textit{Agent-as-models}} \\
    \hline
    GPT-4o & 29.98 & 11.90 & 23.48 & 10.45 & 28.13 & 10.56 & 23.78 & 8.27 & 29.36 & 11.50 & 23.58 & 9.79 & - \\
    Claude-3.7-Sonnet & 28.25 & 9.62 & 22.28 & 7.35 & 32.66 & 9.53 & 27.33 & 8.46 & 29.73 & 9.60 & 23.97 & 7.68 & - \\
    Gemini-2.5-Flash & 32.57 & 11.61 & 26.61 & 10.24 & 31.46 & 9.34 & 23.44 & 8.38 & 30.24 & 10.90 & 25.55 & 9.61 & - \\
    Gemini-2.5-Pro & 45.23 & 21.32 & 39.07 & 16.71 & 44.02 & 19.57 & 35.85 & 13.97 & 44.82 & 20.77 & 37.99 & 15.88 & - \\
    Qwen3-VL-8B
    & 80.40 & 37.10 & 81.00 & 22.94 & 74.58 & 29.94 & 72.07 & 24.14 & 80.33 & 34.96 & 78.08 & 23.30 & -- \\
    \hline
    OS-Atlas-7B-Pro & 49.84 & 24.07 & 35.11 & 12.66 & 47.28 & 22.40 & 37.92 & 15.58 & 48.98 & 23.56 & 36.05 & 13.54 & - \\
    Aguvis-7B & 66.37 & 36.99 & 12.73 & 3.31 & 60.04 & 29.19 & 10.99 & 1.92 & 64.25 & 34.64 & 12.15 & 2.89 & \textbf{15.29} \\
    UI-Venus-Navi-7B & 64.15 & 31.64 & 45.92 & 17.62 & 55.93 & 23.58 & 45.12 & 15.27 & 61.40 & 29.21 & 45.65 & 16.91 & 44.98 \\
    UI-AGILE-7B & 59.30 & 23.66 & 38.22 & 10.92 & 55.99 & 25.73 & 38.10 & 13.74 & 58.19 & 24.29 & 38.13 & 11.77 & 50.32 \\
    AgentCPM-GUI-8B & 71.13 & 38.19 & 42.90 & 16.17 & 71.67 & 37.12 & 43.93 & 18.90 & 71.31 & 37.87 & 43.25 & 16.99 & 37.75 \\
    UI-TARS-7B-DPO & \underline{82.25} & \textbf{55.31} & 49.61 & 19.81 & \underline{79.41} & \textbf{45.80} & \underline{54.48} & \underline{25.23} & \underline{81.30} & \textbf{52.45} & 51.24 & 21.45 & \underline{22.23} \\
    UI-TARS-1.5-7B & 78.92 & 48.60 & \underline{55.74} & \underline{24.13} & 73.83 & 37.35 & 51.27 & 21.32 & 77.22 & 45.21 & \underline{54.21} & \underline{23.28} & 62.47 \\
    AppDeltaAgent-8B (Ours) 
    & \textbf{90.26} & \underline{54.40} & \textbf{83.49} & \textbf{35.39} & \textbf{90.29} & \underline{38.87} & \textbf{81.58} & \textbf{30.71} & \textbf{90.28} & \underline{46.63} & \textbf{82.53} & \textbf{33.05} & 37.12 \\
	\bottomrule
  \end{tabularx}

  \label{tab:static_result}
  \vspace{-3mm}
\end{table*}

\noindent \textbf{Baselines.} For Code2World, we compare against three families of world-model outputs: image-generation systems, code-generation systems, and mixed-generation systems. Image baselines include GPT-Image-2 and Gemini-3.1-Pro-Image; code baselines include GPT-5.4, Claude-Sonnet-4.6, Qwen-family models, MiniMax-M3, and Kimi-K2.6. For action prediction and task execution, we compare general multimodal models and mobile agent frameworks, including Mobile-Agent and AppAgent~\citep{li2024appagent}, as well as GUI-specialized models such as UI-Venus~\citep{gu2025ui} and UI-AGILE~\citep{lian2026ui}.

\noindent \textbf{Metrics.} We follow Code2World and report functional-logic scores $S_{ad}$ and $S_{id}$, visual-quality scores $S_{ele}$ and $S_{lay}$ based on Gemini-3-Flash as judge model,  SigLIP and DINOv2. Static action prediction reports Action Matching Score (AMS), Average Task Progress (ATP), and average output tokens across language and instruction-granularity splits. End-to-end task evaluation reports Success Rate (SR), Progress Rate (PR), False Complete (FC), Overdue Termination (OT), and Unexpected Side Effects (USE). 

\subsection{World-Model Fidelity Evaluation}

\noindent \textbf{Main Results on CMGUIBench-500.}
As shown in Table \ref{tab:cmguibench500-code2world}, AppDeltaWorld achieves the SoTA overall score of 73.51, outperforming all evaluated proprietary multimodal baselines, including GPT-Image-2 and Gemini-3.1-Pro-Image. Compared with the Qwen3-8B base model (50.53), it yields a 22.98 absolute improvement, corresponding to a 45.5\% relative gain. Image-generation models such as GPT-Image-2 preserve global page composition and color fidelity, as reflected by their strong SigLIP and DINOv2 scores, but are less reliable for dense text and fine-grained UI structure: GPT-Image-2 obtains 44.00/41.70 on $S_{ele}/S_{lay}$, compared with 56.26/57.43 for AppDeltaWorld. Nevertheless, its high $S_{ad}/S_{id}$ scores indicate that localized text inconsistencies do not alter the page-level functional state. AppDeltaWorld better balances functional plausibility with element and layout consistency, leading to the strongest overall fidelity. It's important to note that while $S_{lay}$ can assess the relative positions and overall structure of elements within a rendered page, it cannot evaluate whether element coordinates are consistent across the same resolution. Based on manual evaluation, GPT-Image-2 shows better element alignment than AppDeltaWorld.

\noindent \textbf{Ablation Study on AppDeltaWorld.}
To confirm that transition-grounded RAG and diffusion rendering provide complementary benefits, we did the ablation study in Table~\ref{tab:appdeltaworld-ablation}. Removing diffusion decreases the overall score from 73.51 to 70.91, with the largest drops in $S_{ele}$ and $S_{lay}$ (56.26/57.43 to 50.00/50.50), indicating that image synthesis improves visual completeness and fine-grained layout fidelity. Removing RAG causes a larger decline to 67.46 and substantially reduces $S_{ad}/S_{id}$ from 79.69/77.00 to 65.16/69.40, demonstrating the importance of grounded reference retrieval for transition consistency. Removing both components yields the lowest overall (65.68), validating their combined contribution to stable and visually faithful next-screen generation.

\begin{table}[t]
\renewcommand{\arraystretch}{0.9}
\centering
\vspace{-3mm}
\caption{Ablation results of AppDeltaWorld workflow variants on CMGUIBench-500.}
\label{tab:appdeltaworld-ablation}
\scriptsize
\setlength{\tabcolsep}{2.8pt}
\resizebox{\columnwidth}{!}{%
\begin{tabular}{l c c c c c c c}
\toprule
\textbf{Setting} & $S_{ad}$ & $S_{id}$ & $S_{ele}$ & $S_{lay}$ & \textit{SigLIP} & \textit{DINOv2} & \textbf{Overall} \\
\midrule
\textbf{AppDeltaWorld} & 79.69 & 77.00 & \textbf{56.26} & \textbf{57.43} & \textbf{91.84} & \textbf{78.82} & \textbf{73.51} \\
\quad w/o Diffusion & \textbf{81.70} & 76.00 & 50.00 & 50.50 & 90.70 & 76.60 & 70.91 \\
\quad w/o RAG & 65.16 & 69.40 & 50.73 & 51.51 & 90.78 & 77.18 & 67.46 \\
\quad w/o RAG \& Diffusion & 61.40 & 64.60 & 51.10 & 50.70 & 90.40 & 75.90 & 65.68 \\
\bottomrule
\end{tabular}%
}
\vspace{-5mm}
\end{table}

\renewcommand{\arraystretch}{1.3} 

\begin{table*}[t]
\renewcommand{\arraystretch}{1}
  \centering
  \caption{\textbf{Results on MobileGym.} Difficulty SR reports SR within calibrated difficulty strata L1--L4; Diagnostics report False Complete (FC), Overdue Termination (OT), and Unexpected Side Effects (USE). $\pm$ denotes standard deviation across trials.}
  \small
  \setlength{\tabcolsep}{3.0pt}
  \begin{tabularx}{0.9\textwidth}{l*{9}{>{\centering\arraybackslash}X}}
    \toprule
    \multirow{2}{*}{\textbf{Model}}
    & \multicolumn{2}{c}{\textbf{Overall (\%)}}
    & \multicolumn{4}{c}{\textbf{Difficulty SR (\%)}}
    & \multicolumn{3}{c}{\textbf{Diagnostics (\%)}} \\
    \cmidrule(r){2-3}
    \cmidrule(r){4-7}
    \cmidrule(r){8-10}
    & \textbf{SR} & \textbf{PR} & \textbf{L1 (20)} & \textbf{L2 (73)} & \textbf{L3 (83)} & \textbf{L4 (80)} & \textbf{FC} & \textbf{OT} & \textbf{USE} \\
    \hline
    \multicolumn{10}{c}{\textit{Proprietary models}} \\
    \hline
    Gemini 3.1 Pro & 58.8 {\scriptsize $\pm$1.4} & 72.1 & 97.5 & 83.6 & 63.3 & 21.9 & 34.0 & 0.2 & 5.5 \\
    Doubao-Seed-2.0-Pro & 52.0$^\dagger$ & 63.6 & 100.0 & 93.2 & 48.2 & 6.2 & 33.6 & 0.4 & 4.7 \\
    Qwen3.6-Plus & 45.7$^\dagger$ & 59.2 & 100.0 & 78.1 & 44.6 & 3.8 & 34.0 & 0.4 & 14.5 \\
    GPT-5.5 & 36.1$^\dagger$ & 51 & 85 & 49.3 & 34.1 & 13.8 & 30.9 & 0.8 & 19.9 \\
    \hline
    \multicolumn{10}{c}{\textit{Open-source generalist models}} \\
    \hline
    Qwen3-VL-8B-Instruct & 10.2 {\scriptsize $\pm$0.8} & 22.0 & 66.2 & 13.4 & 3.6 & 0.0 & 14.1 & 1.2 & 5.6 \\
    Qwen3-VL-4B-Instruct & 9.4 {\scriptsize $\pm$0.6} & 20.1 & 71.2 & 12.3 & 0.6 & 0.3 & 15.9 & 0.4 & 10.0 \\
    \hline
    \multicolumn{10}{c}{\textit{Open-source GUI-specialized models}} \\
    \hline
    AutoGLM-Phone-9B & 20.0 {\scriptsize $\pm$1.3} & 35.3 & 86.2 & 33.6 & 9.6 & 1.9 & 39.6 & 0.6 & 12.6 \\
    UI-Venus-1.5-8B & 15.4 {\scriptsize $\pm$2.4} & 28.3 & 85.0 & 21.9 & 6.0 & 1.9 & 22.9 & 0.5 & 7.7 \\
    GUI-Owl-1.5-8B-Think & 15.1 {\scriptsize $\pm$0.9} & 28.8 & 76.2 & 26.0 & 4.2 & 1.2 & 30.4 & 0.9 & 14.1 \\
    UI-TARS-1.5-8B & 13.8 {\scriptsize $\pm$1.7} & 26.3 & 77.5 & 21.9 & 3.0 & 1.6 & 38.6 & 0.2 & 11.0 \\
    Step-GUI-4B & 12.9 {\scriptsize $\pm$1.1} & 25.7 & 83.8 & 17.8 & 2.4 & 1.6 & 37.0 & 0.8 & 7.6 \\
    \midrule
    AppDeltaAgent-8B (Ours) & 14.1 {\scriptsize $\pm$1.4} & 26.4 & 83.8 & 19.5 & 5.1 & 0.9 & 36.0 & 0.7 & 19.7 \\
    \bottomrule
  \end{tabularx}
  \label{tab:main-results}
  \vspace{-3mm}
\end{table*}
\subsection{Agent Policy Evaluation}

\noindent \textbf{AndroidLens.}
As shown in Table~\ref{tab:static_result}, AppDeltaAgent-8B achieves the highest AMS across all language and instruction-granularity splits, with particularly strong gains in task progress under high-level instructions. Relative to the Qwen3-VL-8B base model, it improves Total-LL AMS/ATP from 80.33/34.96 to 90.28/46.63, corresponding to a 33.4\% relative ATP gain, and improves Total-HL AMS/ATP from 78.08/23.30 to 82.53/33.05, yielding a 41.8\% relative ATP gain. These improvements are consistent across both Chinese and English tasks, indicating that world-model-generated supervision enhances not only local action matching but also sustained progress from abstract task descriptions. Its larger advantage on high-level instructions further suggests that action-conditioned successor states provide useful supervision for long-horizon decision making beyond static grounding alone. These results further demonstrate that, after quality filtering, the experience provided by AppDeltaWorld can be used as a substitute for real-world environments.

\begin{figure}[t]
    \centering
    \includegraphics[width=\columnwidth]{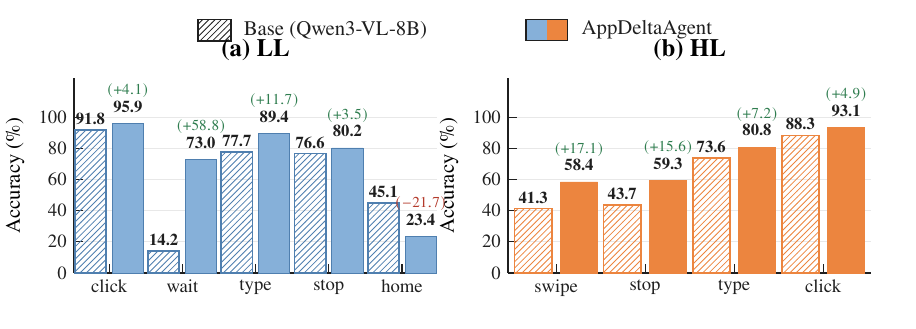}
    \vspace{-0.5cm}
    \caption{Per-action-type accuracy of Qwen3-VL-8B and AppDeltaAgent on AndroidLens. }
    \label{fig:action-type-accuracy}
    \vspace{-0.5cm}
\end{figure}

\noindent\textbf{MobileGym.}
As shown in Table~\ref{tab:main-results}, AppDeltaAgent-8B achieves an SR of 14.1, representing a relative improvement of 38.2\% over the baseline model Qwen3-VL-8B-Instruct (10.2), while increasing PR from 22.0 to 26.4. Among open-source GUI agents, AppDeltaAgent surpasses several competitive 8B baselines, though a substantial gap remains to larger-scale open-source and proprietary models. The improvements are most pronounced on L1 (66.2$\rightarrow$83.8) and L2 (13.4$\rightarrow$19.5) tasks. These results demonstrate that hybrid SFT training not only improves local action prediction but also enhances execution capability during long-horizon GUI interactions. Unlike conventional demonstration data that primarily provides static behavioral supervision, AppDeltaWorld rollouts introduce additional interaction trajectories guided by state transition constraints, complementing public GUI datasets with closed-loop execution experiences. 

\noindent\textbf{MobileWorld.}
To further evaluate the effectiveness in realistic application scenarios, we test AppDeltaAgent on MobileWorld. Unlike the controlled simulated environment in MobileGym, MobileWorld directly executes agents on real Android applications, introducing more complex application behaviors and challenges arising from longer interaction dependencies. As shown in Table~\ref{tab:mobileworld-results}, AppDeltaAgent-8B achieves a GUI-only SR of 14.9, improving Qwen3-VL-8B-Instruct (9.4) by 58.5\%, while increasing the average number of interaction steps from 24.8 to 30.1.In this environment that better reflects real-world usage scenarios, AppDeltaAgent continues to achieve consistent improvements, demonstrating that the benefits of hybrid SFT training are not limited to controlled simulated environments. AppDeltaWorld rollouts are not intended to replace existing GUI demonstration data; instead, they provide additional interaction supervision constrained by state transitions, thereby enhancing the agent's ability to maintain coherent interaction trajectories. These results suggest that augmenting offline demonstration data with AppDeltaWorld rollouts can improve GUI task execution in practical environments, yielding benefits beyond controlled simulation settings.

\begin{table}[t]
\renewcommand{\arraystretch}{0.9}
\centering
\vspace{-3mm}
\caption{\textbf{MobileWorld evaluation results.} We report the GUI-only success rate and average steps.}
\label{tab:mobileworld-results}
\scriptsize
\setlength{\tabcolsep}{3.2pt}
\resizebox{0.9\columnwidth}{!}{%
\begin{tabular}{l c c c}
\toprule
\textbf{Model} & \textbf{Overall SR} & \textbf{GUI-only SR} & \textbf{Steps} \\
\midrule
Claude-4.5-Sonnet + UI-Ins-7B & 43.8 & 47.8 & 26.6 \\
Gemini-3-Pro + UI-Ins-7B & 46.3 & 55.6 & 24.2 \\
GPT-5 + UI-Ins-7B & 51.7 & 54.0 & 27.8 \\
\midrule
Doubao-1.5-UI-TARS & 20.9 & 26.3 & 20.9 \\
GUI-Owl-7B & 4.5 & 7.7 & 20.6 \\
GUI-Owl-32B & 5.5 & 8.5 & 24.0 \\
UI-Venus-7B & 5.5 & 8.5 & 26.7 \\
UI-Venus-72B & 10.4 & 16.4 & 34.2 \\
Qwen3-VL-8B & 5.5 & 9.4 & 24.8 \\
Qwen3-VL-32B & 9.0 & 11.9 & 27.1 \\
Qwen3-VL-235B-A22B & 9.5 & 12.8 & 26.9 \\
\midrule
AppDeltaAgent-8B (Ours) & -- & 14.9 & 30.1 \\
\bottomrule
\end{tabular}%
}
\vspace{-5mm}
\end{table}

\begin{figure}[t]
    \centering
    \includegraphics[width=\columnwidth]{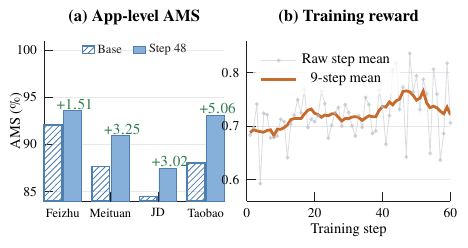}
    \vspace{-0.5cm}
    \caption{World-model-based test-time RL results on AndroidLens. (a) AMS before and after 48 steps for four improved apps. (b)~Mean reward over the first 60 steps.}
    \label{fig:ttrl-analysis}
    \vspace{-0.5cm}
\end{figure}

\subsection{World-Model-Based RL-Training}

\noindent \textbf{World-Model-Based Self-score RL-Training.} As \citet{xu2026mobile} explained, world models can help agents determine and preview the expected result of the current action. We designed a self-score RL experiment: for each current-screen and instruction pair, the policy rollouts eight actions, and AppDeltaWorld predicts and renders the next state. The policy model then self-scores the sampled action, evaluating action consistency and instruction progress to produce the reward signal. Figure~\ref{fig:ttrl-analysis}(a) shows that the Step-48 policy improves AMS by 1.51, 3.25, 3.02, and 5.06 points on Feizhu, Meituan, JD, and Taobao, respectively. Figure~\ref{fig:ttrl-analysis}(b) further shows that, despite substantial per-update variation from policy sampling and world-model rendering, the reward rises overall during the first 60 updates from 68 to 76. These results suggest that AppDeltaWorld can support app-specific policy adaptation at test time while avoiding additional interaction with real apps.

\begin{figure}[t]
    \centering
    \includegraphics[width=\columnwidth]{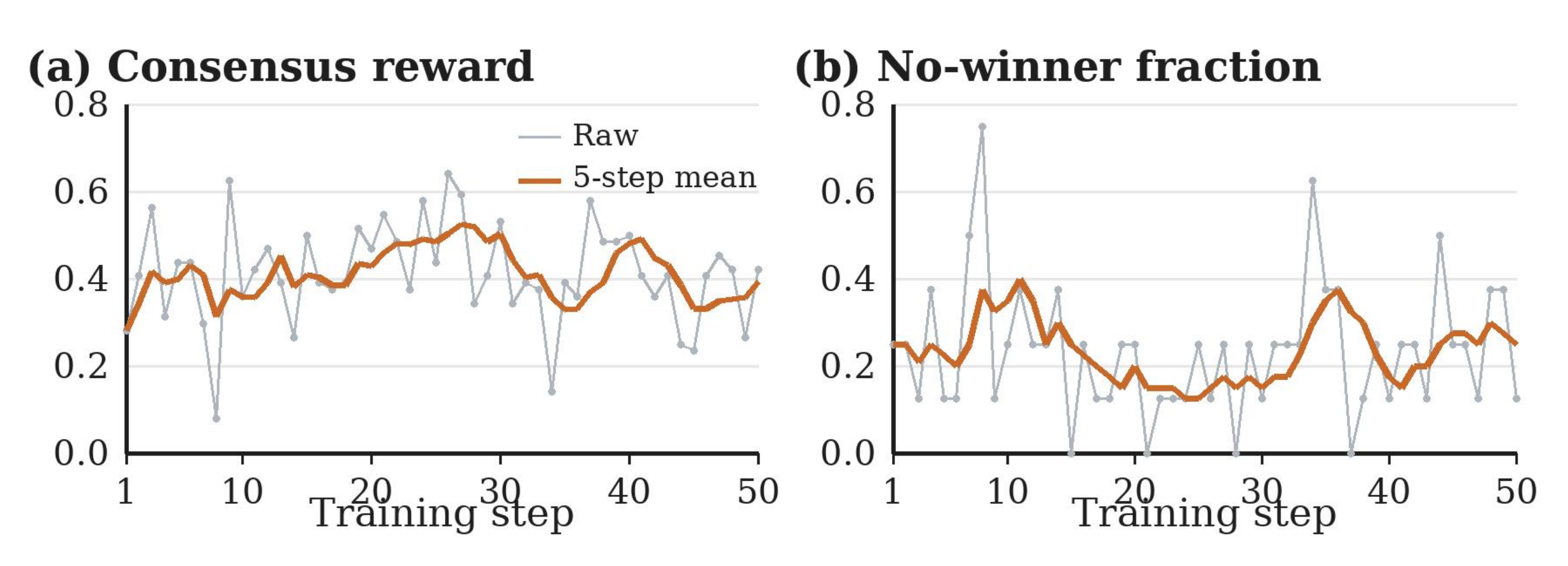}
    \vspace{-0.4cm}
    \caption{Training statistics for world-model-based consensus-reward RL on AndroidLens low-level tasks. (a)~Mean consensus reward. (b)~Fraction of groups without a unique supported winner. }
    \label{fig:ttrl-consensus}
    \vspace{-0.4cm}
\end{figure}

\noindent \textbf{World-Model-Based Clustering reward RL-Training.}
We further study whether AppDeltaWorld can provide a reward without an LLM judge or action labels. Following the voting principle of test-time RL~\citep{zuo2025ttrl}, each update samples eight AndroidLens instructions and eight action rollouts per instruction (64 rollouts in total). AppDeltaWorld predicts the successor state for every rollout. Within each eight-rollout group, we cluster predicted states using a 0.60 visual-similarity and 0.40 text-similarity score, with threshold 0.82. A rollout receives reward 1 only if its predicted state belongs to the unique largest cluster with at least two members; outliers, invalid rollouts, and tied or unsupported clusters receive reward 0. Thus, the signal rewards action outcomes that are independently supported by multiple rollouts, without requiring model-based judging or ground-truth outcomes.
We adapt AppDeltaAgent with single-step GRPO on AndroidLens low-level tasks and analyze the first 50 completed updates. Figure~\ref{fig:ttrl-consensus} shows that the consensus reward averages 0.412 and is typically between 0.4 and 0.5. The average winning-cluster support is only 3.295 of 8 rollouts, indicating that the selected outcome is usually supported by three or four rollouts rather than a strong majority. Moreover, 23.5\% of rollout groups have no unique supported winner on average, with the fraction reaching 0.4 or higher in some updates. These statistics reveal a limitation of the current dense visual--text state representation: semantically equivalent successor states are not yet sufficiently consistent to form reliable clusters. Better state-consistency metrics or learned aggregation mechanisms are therefore needed before consensus can provide a strong standalone reward. 

\subsection{Analysis Experiments}

We further analyze where AppDeltaAgent improves over Qwen3-VL-8B, and how synthetic rollout quantity affects training. The first question concerns whether the AndroidLens gains come from stronger coordinate grounding or from better action-type and long-horizon planning. The second question measures how performance scales when AppDeltaWorld rollout data is added to the public SFT mixture.

\noindent\textbf{Action-Type Improvements Analysis.}
Figure~\ref{fig:action-type-accuracy} shows that AppDeltaAgent's advantage over Qwen3-VL-8B is not driven by coordinate localization. On low-level instructions, the largest gains come from tool-oriented actions: wait improves from 14.19 to 72.97, type from 77.66 to 89.37, but click only from 91.82 to 95.88. On high-level instructions, the gains shift toward exploration and termination: swipe rises from 41.31 to 58.37, stop from 43.73 to 59.32. Together, these results attribute the LL gains mainly to more accurate action-type decisions and text entry, and the HL gains to better page exploration, navigation progress, and stop timing, rather than to improved localization of page elements. This finding suggests that, even when page-element positions do not perfectly align with those in real apps, the higher-level interaction experience remains valuable for policy learning.

\begin{figure}[t]
    \centering
    \includegraphics[width=\columnwidth]{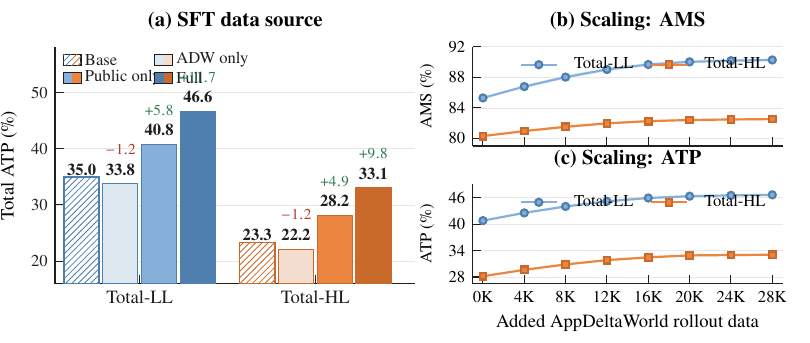}
    \vspace{-0.5cm}
    \caption{SFT data ablation and AppDeltaWorld data scaling on AndroidLens. (a)~Total-LL/HL ATP under different SFT mixtures. (b)--(c)~AMS and ATP when adding AppDeltaWorld rollout data in 4K increments.}
    \label{fig:sft-data-ablation-scaling}
    \vspace{-0.6cm}
\end{figure}

\noindent\textbf{SFT Data Scaling.}
Figure~\ref{fig:sft-data-ablation-scaling}(a) shows that public SFT data provide the foundation for action-model training: Public only improves Total-LL/HL ATP by 5.83/4.87 over the Qwen3-VL-8B base. AppDeltaWorld rollouts provide complementary supervision that further improves generalization, raising Total-LL ATP from 40.79 to 46.63 and Total-HL ATP from 28.17 to 33.05 when added to the public data. However, AppDeltaWorld data cannot be used effectively in isolation because the world model still exhibits biases in fundamental interaction patterns; under the ADW-only setting, Total-LL/HL ATP instead decreases by 1.16/1.15 relative to the base. Figures~\ref{fig:sft-data-ablation-scaling}(b)--(c) further examine the effect of mixing different amounts of AppDeltaWorld data with public supervision. As the amount of AppDeltaWorld data increases from 0K to 28K, all four metrics improve monotonically but with diminishing returns: approximately 75\% of the final gain is achieved by 12K, and performance saturates after 20K. This trend indicates that the diversity of action-model supervision remains bounded by the world model's experience and cannot be expanded indefinitely through instruction augmentation.

\section{Conclusion}
We introduced AppDeltaWorld, a transition-grounded delta code world model that combines action-constrained hierarchical HTML retrieval, executable next-screen generation, visual-asset synthesis, and world-model-in-the-loop rollout construction. AppDeltaWorld achieves the strongest overall fidelity on CMGUIBench-500, while its generated trajectories consistently improve AppDeltaAgent on AndroidLens, MobileGym, and MobileWorld. These results demonstrate that a transition-grounded GUI world model can serve as a scalable complement to real interaction environments for generating useful policy supervision.
\clearpage
\newpage
\bibliography{aaai2027}

\end{document}